\documentclass[letterpaper,journal]{IEEEtran}
\usepackage{amsmath,amsfonts}
\usepackage{algorithmic}
\usepackage{algorithm}
\usepackage{array}
\usepackage[caption=false,font=normalsize,labelfont=sf,textfont=sf]{subfig}
\usepackage{textcomp}
\usepackage{stfloats}
\usepackage{url}
\usepackage{verbatim}
\usepackage{graphicx}
\usepackage{multirow}
\usepackage{pifont}
\usepackage{array}
\usepackage{amsthm}
\usepackage{booktabs}
\usepackage{tabularx}
\usepackage{cite}
\usepackage{xcolor}
\DeclareRobustCommand{\rev}[1]{#1}
\usepackage{amsthm}
\usepackage{makecell}
\newtheorem{definition}{Definition}
\theoremstyle{plain}
\newtheorem{lemma}{Lemma}[section]
\newtheorem{theorem}{Theorem}
\begin{document}

\title{Enhancing Fatigue Detection through Heterogeneous Multi-Source Data Integration and Cross-Domain Modality Imputation}

\author{
Luobin Cui,
Yanlai Wu,
Tang Ying,~\IEEEmembership{Fellow,~IEEE,} and
Weikai Li
\thanks{L. Cui and Y. Wu are with the Department of Electrical and Computer Engineering, Rowan University, Glassboro, NJ, 08028 USA (e-mail: cuiluo77@students.rowan.edu, wuyanl37@rowan.edu).}
\thanks{Ying Tang (the corresponding author) is with the Department of Electrical and Computer Engineering, Rowan University, Glassboro, NJ, 08028 USA  (e-mail: tang@rowan.edu).}
\thanks{Weikai Li is with the School of Computer and Artificial Intelligence,
Shandong Jianzhu University (e-mail: leeweikai@outlook.com).}
}

\IEEEpubid{}

\maketitle

\begin{center}
\footnotesize This work has been submitted to the IEEE for possible publication. Copyright may be transferred without notice, after which this version may no longer be accessible.
\end{center}

\begin{abstract}
Fatigue detection for human operators is important in safety-related applications such as aviation, mining, and long-haul transport. Reliable estimation of operator fatigue can support timely warnings, adaptive task scheduling, takeover reminders, and other safety-management decisions in human-machine systems. However, the effectiveness of these functions depends on whether fatigue-related signals can be reliably captured in the deployment environment. While many studies have shown the value of high-fidelity sensors in controlled laboratory environments, their performance often degrades when used in real-world settings because of noise, lighting conditions, and field-of-view constraints, thereby limiting their practical use. This paper formalizes a deployment-oriented setting for real-world fatigue detection, where high-quality sensors are often unavailable in practical applications. To address this issue, we use knowledge from heterogeneous source domains, including high-fidelity sensors that are difficult to deploy in the field but commonly used in controlled environments, to assist fatigue detection in the real-world target domain. Based on this idea, we design a heterogeneous and multi-source fatigue-detection framework that uses the available modalities in the target domain while leveraging diverse configurations in the source domains through cross-domain modality imputation based on shared modalities. As a proof-of-concept, we intentionally employ well-established algorithms for both knowledge-transfer and fatigue-detection components to test the framework's core idea. Experiments using a field-deployed sensor setup and two publicly available human-fatigue datasets demonstrate the feasibility of the proposed framework in sensor-constrained scenarios and show that knowledge from heterogeneous source domains can support fatigue detection when high-fidelity modalities are unavailable in the target domain.

\end{abstract}

\begin{IEEEkeywords}
fatigue detection, cross-domain modality imputation, human-machine systems, computational behavior modeling, domain adaptation, multimodal learning
\end{IEEEkeywords}

\section{Introduction}
Physiological responses and behavioral patterns of individuals under fatigue provide rich information to understand the impacts of jeopardized human capacity, from which proper mitigation control can be determined. In practical human-machine systems, such information can also be used as an input to warning, scheduling, intervention, and safety-management decisions. For these functions to be reliable, fatigue-related data must be captured consistently in the deployment environment. However, the complexity and difficulty of data acquisition present several limitations. While recent advances in sensor technology have improved objective fatigue detection, concerns about sensor reliability and accessibility remain \cite{Hooda2022}.

Unreliable sensors can result in data loss, compromised detection accuracy, or, in the worst case, complete malfunction. To address this issue, recent efforts have explored modality imputation strategies to augment incomplete or noisy datasets. For instance, sensor-based data imputation \cite{zhang2025differentiable} has emerged as an effective approach. Bird et al. (2021) showed that GPT-2 can generate synthetic electroencephalography (EEG) and electromyography (EMG), which, when combined with real training data, improved classification accuracy \cite{9345373}. Similarly, \cite{10178849} employed generative adversarial networks (GANs) to enhance data for motor-imagery classification in brain-computer interface systems. Along this direction, \cite{zeng2021eeg} proposed Generative-DANN (GDANN), a hybrid model that integrates GANs with a Domain-Adversarial Neural Network (DANN) architecture to address cross-subject EEG variability in fatigue detection. By aligning feature distributions across subjects, GDANN improved generalizability and achieved high accuracy in cross-subject fatigue classification tasks. When performing modality imputation for fatigue detection, the process must consider inherent correlations among physiological modalities to collaboratively determine fatigue. To that end, RAINDROP \cite{zhang2021graph} addressed this by using graph neural networks to model these correlations in irregularly sampled multivariate time-series data.

Although the aforementioned approaches address the challenge of sensor unreliability through data imputation, few efforts have tackled the related issue of sensor accessibility. In many real-world scenarios, high-end sensors, such as EEG, imaging systems, and pupil-dilation monitors, while reliable and accurate, are often expensive. The extensive calibration, routine maintenance, and specialized operator training required make these sensors difficult to operate at scale in real-world deployments. Moreover, these sensors are highly sensitive to environmental factors such as noise, lighting conditions, and field-of-view constraints. These sensitivities complicate data collection and analysis in real-world scenarios, which can degrade model accuracy and robustness \cite{e19080385}. As a result, studies involving these sensors are often confined to simulated and controlled environments. For example, high-resolution EEG was deployed in a simulated environment, together with heart rate and eye-blink rate, to analyze mental fatigue or drowsiness during car driving \cite{6347469}. However, such simulation-based setups often fail to capture the complexity and uncertainty of actual driving conditions \cite{8482470}, which in turn limits the generalizability of models trained in these idealized environments. Nonetheless, transferring knowledge gained from controlled settings remains useful, provided that it is adapted appropriately, thereby offering a basis for practical fatigue detection in real-world scenarios where direct sensor deployment is constrained. Note that this type of transfer requires a form of cross-domain adaptation \cite{jimenez2025mmda} that goes beyond the capabilities of existing generative methods. Current approaches, such as the GAN- and GPT-based models mentioned earlier, have proven effective for within-domain synthesis, where additional data are generated for modalities already present at both training and deployment (e.g., generating more EEG from partial EEG). However, they are not designed for cross-domain knowledge transfer in settings where an entire modality is absent from the target domain during both training and deployment, for example, when EEG-related knowledge is leveraged although no EEG has ever been observed in the target system. This limitation highlights a gap: the need for methods that can support cross-domain modality imputation to address real-world sensor accessibility constraints.

Based on these observations, this paper proposes a cross-domain fatigue detection framework that enables real-world fatigue detection systems to learn from and be enhanced by controlled-environment data, even when the data originate from different sensor sources. The intuition is that deployable physiological and behavioral modalities, such as HR, PPG, and ACC, reflect responses to the same underlying fatigue state as high-fidelity but less deployable modalities observed in controlled environments, such as EEG. When such deployable modalities are shared across source and target domains, they provide an overlapping feature space through which the model can learn source-domain relationships between deployable and less deployable modalities and transfer these relationships to the target domain where the latter are unavailable. In this way, the proposed approach extends the utility of high-fidelity but deployment-constrained sensors to practical settings where they are otherwise unavailable. This work serves as a proof-of-concept, aiming to demonstrate the feasibility of the proposed heterogeneous and multi-source fatigue-detection framework. To isolate the
contribution of the framework design itself, we intentionally
employ widely used, well-understood models for both the
knowledge transfer and fatigue detection components, emphasizing
that the innovation lies in their integration rather than in
the specific algorithms themselves. Experiments are then conducted in the proposed sensor-constrained setting to evaluate whether the integrated framework can support fatigue detection when high-fidelity modalities are unavailable in the target domain. The proposed framework is presented with the following key contributions:

\begin{itemize}
    \item A practical yet underexplored problem setting for real-world fatigue detection is formally defined. This setting reflects the need to operate fatigue-detection systems with context-appropriate and accessible sensors in the field while leveraging knowledge from high-fidelity but deployment-constrained source modalities.
    \item A heterogeneous multi-source fatigue-detection framework is proposed as a solution to this problem. Within this framework, shared physiological and behavioral modalities are used as the common input space for learning source-domain modality imputers, estimating less deployable modalities absent from the target domain, and integrating available and imputed information for target-domain fatigue prediction.
    \item The framework is evaluated using a practical target-domain sensor configuration and two auxiliary public datasets. The results demonstrate the feasibility of transferring fatigue-relevant information across differently instrumented domains, with observed improvements in target-domain fatigue detection under sensor-constrained conditions.
\end{itemize}

\section{Related Work}

This section reviews key developments in fatigue assessment. Section II.A examines existing fatigue-detection methods, highlighting their advantages and limitations, which underscore the need for a new approach that builds on their strengths while addressing their shortcomings. Section II.B explores current data synthesis and modality-imputation methods for mitigating data scarcity, along with their constraints in practical applications. Finally, Section II.C discusses the use of open datasets for fatigue assessment and the associated challenges related to sensor discrepancies.

\subsection{Fatigue Assessment}\label{subsec:traditional}

Traditional fatigue assessment methods are based on subjective evaluation tools, including self-reports and questionnaires. Since the last century, many classic fatigue assessment scales have been used, including the Stanford Sleepiness Scale (SSS) \cite{Shahid2012SSS}, proposed by Stanford University; the Multidimensional Fatigue Inventory (MFI) \cite{smets1995multidimensional}, developed by Smets et al.; and the International Fitness Scale (IFIS) \cite{Ortega2011}, which measures changes in overall vigor and affect. These scales cover multiple dimensions of fatigue and serve as the basis for traditional fatigue assessment. However, both questionnaires and self-reports are affected by an individual's psychological state and external environment, resulting in subjective and inconsistent outcomes \cite{gawron2016overview}. Moreover, most of these methods \cite{brown2016fatigue,jap2009using,kar2010eeg} are post hoc assessments, making them unsuitable for real-time fatigue monitoring and preventive measures, both of which are needed in high-stakes environments, such as driving \cite{app14073016} and aviation \cite{article111,10552405}.

To address these limitations, researchers have explored more objective measures of human fatigue. In fact, a person's fatigue state can be reflected in changes in physiological signals, such as heart rate (HR), heart rate variability (HRV), electrocardiogram (ECG), EEG \cite{9464716,10488090}, galvanic skin response (GSR), skin temperature (ST), blood pressure (BP), EMG \cite{7516656,9551478}, and blood oxygen saturation (SpO2). Studies have shown that when drivers enter a fatigued state, their HRV metrics change clearly. For example, a decrease is often observed in time-domain measures, such as natural variability in interbeat intervals and differences between consecutive heartbeats. This suggests that the ability of the body to regulate heart activity is weakened. Similarly, HRV indicators in the frequency domain, such as the low-frequency and high-frequency ratios, often become unstable, reflecting an imbalance in autonomic regulation \cite{7473750}. GSR signals during fatigue are also usually characterized by a rise in baseline conductance and response fluctuations, which are closely correlated with enhanced sympathetic nerve activity \cite{Markiewicz2022}. Through real-time monitoring and joint analysis of these indicators, subtle changes in the driver's physiological state can be captured more precisely, thereby improving the sensitivity and accuracy of the fatigue-detection system \cite{10324398}. With continued technological advancement, physiological signal-acquisition devices are becoming smaller and more user-friendly, enabling real-time fatigue monitoring \cite{mohanavelu2017assessment,hu2024physiological}.

Among various physiological signals, EEG has gained attention in fatigue detection due to its ability to directly measure brain activity with high accuracy \cite{cheng2022cognitive}. However, traditional EEG systems are expensive, require an intrusive setup and specialized personnel for operation, making real-world deployment challenging. As a result, the majority of EEG-based fatigue studies remain confined to controlled laboratory settings \cite{trejo2015eeg,cheng2011mental,8982166,10107629}. Although portable EEG devices have emerged for field applications, their data quality is lower than that of laboratory-grade devices. Additionally, they are susceptible to external noise and interference, affecting their stability and detection accuracy \cite{larocco2020systemic,shashidhar2020real}.

In response to these challenges, researchers have explored noninvasive alternatives, leading to the rise of vision-based fatigue detection \cite{LIU2022105399}. These methods assess fatigue by analyzing facial features, such as eyelid-closure time \cite{10054532}, yawning frequency \cite{8808931}, and body movement \cite{10260460,9926591}, from images captured by cameras in real time. Despite their many advantages, these methods also face challenges, such as lighting variations and camera angles, which can affect image quality and ultimately detection accuracy. In addition, long-term monitoring of human subjects' facial and body movements in work environments raises privacy concerns \cite{Shen2024} and may lead to employee resistance. Along with this trend, alternative visual signals, such as pupil dilation and gaze tracking, have gained attention in the field of fatigue detection. The eyes are often considered the most intuitive window into mental state, as they can directly reflect the immediate response to fatigue and provide high accuracy in fatigue assessment. Studies have shown that pupil diameter and gaze focus are important indicators for assessing workload and fatigue status \cite{krejtz2018eye,doi:10.1177/21695067231192895,doi:10.1076/0271-3683(200007)2111-ZFT535}, including our own study \cite{10799435}, which collected eye and physiological data for combined fatigue detection. However, while pupil dilation and gaze tracking offer promising insights, they are also sensitive to external environmental factors. Pupil dilation can be influenced by ambient lighting conditions, which may reduce its reliability in dynamic or uncontrolled environments. Similarly, gaze tracking can be affected by head movements, occlusions, and variations in camera positioning, limiting its stability and accuracy unless the camera is confined to a controlled environment, such as the HoloLens setup used in our study \cite{10799435}.

In summary, different sensor signals possess distinct advantages in fatigue detection, with some proving more effective than others. However, their effectiveness is highly contingent on application context and deployment setting. For example, EEG performs well in controlled settings but can be challenging to deploy in real-world contexts because of its susceptibility to noise and constraints on setup and usability. Although commercial wearable EEG systems exist, practical use outside the lab often requires additional artifact handling and user training. Similar sensitivity issues rule out the use of pupil dilation in practical applications and require the use of gaze tracking within a confined setup. Therefore, researchers usually select available and/or context-appropriate resources that best align with their research objectives. Even so, we should not overlook the value of knowledge learned in settings where sensors that are impractical in the target context but effective in controlled settings have been effectively applied. If a mechanism exists to transfer that knowledge, it could be used to augment fatigue detection in environments where those sensors are unavailable or impractical, thereby preserving their contribution without requiring their direct deployment.

\subsection{Data Synthesis \rev {and Modality Imputation} }\label{subsec:data}

Data synthesis have demonstrated significant advantages in dealing with data scarcity and incompleteness in recent years. The former generates artificial data while preserving the statistical or structural characteristics of real datasets, whereas the latter estimates missing information from observed signals. These strategies are  particularly relevant in physiological signal analysis, where data collection can be challenging because of factors such as sensor availability, sensitivity to noise, and subject-related variability, leading to widespread adoption in this domain \cite{NEIFAR2025103127}.

Early developments in the are primarily leveraged statistical regression techniques, either to construct additional synthetic data through parameterized modeling of observed data distributions or to fill missing values from related observed variables. As datasets grew in complexity and dimensionality, regularized models such as ridge and lasso were introduced to improve generalization. However, despite these advancements, statistical models remain fundamentally limited in their ability to capture long-term dependencies and nonlinear temporal dynamics prevalent in complex physiological signals {\cite{vetter2024patterns}}.

Deep learning, on the other hand, has provided effective solutions to this challenge. Recurrent Neural Networks (RNNs), particularly Long Short-Term Memory (LSTM) networks, have proven effective in capturing long-term temporal dependencies in multivariate time-series data. Several studies have applied these models in a variety of applications, such as GRU-D, which incorporates decay mechanisms and masking for clinical time-series data \cite{che2016recurrent}; an LSTM-based imputation model for ECG signals that outperforms classical methods \cite{verma2019accurate}; and BRITS, which enables bidirectional gradient-based imputation without strong distributional assumptions \cite{cao2018brits}. In parallel, Convolutional Neural Networks (CNNs) have shown good performance in extracting local spatial and temporal features. Multimodal physiological-signal fusion has also been studied through correlation-based fusion strategies \cite{shen2024tensor}. The strength of CNNs has inspired hybrid models that combine CNNs with LSTMs to jointly use spatial and temporal dependencies for more accurate imputation and signal reconstruction \cite{Eum2022Imputation}. Additionally, GANs have emerged as a common approach for producing high-fidelity synthetic signals that are statistically similar to real ones. Examples can be found in \cite{10239214,NIA2024110129}.

Despite this progress, current data-synthesis and modality-imputation techniques across both classical and deep-learning paradigms typically operate under a critical limitation: they rely solely on observed data from a single modality or domain to synthesize or impute new data of the same type. This assumption restricts their generalizability and limits their ability to model cross-domain relationships or estimate unseen modalities in the target domain. This limitation is especially problematic in real-world physiological settings, where data are not only sparse and incomplete, but also highly heterogeneous across sensing devices, sensing modalities, and subject populations. 



\subsection{Open Fatigue Detection Datasets}\label{subsec:datasets}

A review of relevant literature has identified several open datasets for fatigue detection. As shown in Table~\ref{dataset}, the use of diverse physiological signals reaffirms that researchers tend to select sensors that are readily accessible and convenient. Meanwhile, certain signals recur across datasets, including HR, PPG, GSR, acceleration (ACC), and ST.

The commonality and diversity across datasets present both challenges and opportunities. For example, while FatigueSet \cite{10.1007/978-3-030-99194-4_14}, MEFAR \cite{DERDIYOK2024109896}, and VPFD \cite{10799435} share similar data types, they use different sensors—Empatica E4 for the first two datasets and Google Pixel Watch~2 for the third. These variations in data collection methods, sensor specifications, and pre-processing can hinder standardization and generalization.

At the same time, these three datasets are suitable for cross-dataset exploration because they all provide fatigue annotations based on self-reported binary fatigue scores. The only distinction lies in FatigueSet, which separates mental and physical fatigue, whereas MEFAR and VPFD do not. For consistency across datasets, the mental- and physical-fatigue annotations are combined into a unified label: a sample is treated as positive if either its mental- or physical-fatigue annotation indicates fatigue. This keeps the binary target definition consistent with MEFAR and VPFD. 

Beyond differences in sensing devices, the datasets also differ in modality availability.
FatigueSet \cite{10.1007/978-3-030-99194-4_14} and MEFAR \cite{DERDIYOK2024109896}, collected under controlled laboratory conditions, include high-fidelity modalities—such as EEG in both datasets and ECG in FatigueSet—that are often impractical for real-world deployment. In contrast, the VPFD \cite{10799435} dataset was designed for real-world use and thus relies on more accessible modalities, without EEG and ECG. This distinction directly motivates our evaluation of the central hypothesis posed in the Introduction: Can fatigue-relevant knowledge learned from high-resolution EEG and ECG sources be effectively transferred to enhance fatigue prediction in settings like VPFD, where only context-appropriate modalities are available?

This question positions our work differently from both same-domain data imputation and standard cross-domain fatigue recognition. Same-domain data imputation typically reconstructs data within the same modality space or the same domain, whereas our work focuses on cross-domain modality imputation: shared deployable modalities are used to estimate less deployable modalities that are unavailable in the target domain. Although the imputed modalities are used to support downstream fatigue detection, this differs from standard cross-domain fatigue recognition, which usually adapts fatigue classifiers across domains that share the same feature space. Our work instead addresses a heterogeneous sensor-constrained setting in which source and target domains may contain different modality sets, and shared modalities are used, with the support of domain alignment, to transfer cross-modal fatigue-relevant information to the target domain.

\begin{table*}[htbp]
  \centering
  \caption{Summary of fatigue detection datasets and acquisition devices}
  \label{dataset}
  \resizebox{\textwidth}{!}{
     \begin{tabular}{|>{\centering\arraybackslash}m{2.7cm}|c|c|c|c|c|c|c|c|c|c|c|c|c|}
      \hline
      \multirow{2}{*}{{Dataset}} & \multirow{2}{*}{{Human Subjects}} & \multirow{2}{*}{{Hours}} & \multicolumn{11}{c|}{{Data Types}} \\
      \cline{4-14}
      & & & {PPG} & {GSR} & {HR} & {ST} & {ACC} & {EYE} & {EEG} & {ECG} & {EMG} & {BP} & {FACE} \\
      \hline
      CogBeacon \cite{technologies7020046} & 19 & 35  &  &  &  &  &  & \checkmark &  &  &  &  & \checkmark \\
      \hline
      CLAS \cite{8967457}                 & 62 & 31  & \checkmark & \checkmark & \checkmark &  &  &  &  & \checkmark &  &  &  \\
      \hline
      MePhy \cite{Gabbi2024UnderstandingFT} & 60 & 8   &  & \checkmark &  &  &  & \checkmark &  & \checkmark & \checkmark &  & \checkmark \\
      \hline
      OperEYEVP \cite{s23136197}           & 10 & 10  &  &  & \checkmark &  &  & \checkmark &  &  &  & \checkmark & \checkmark \\
      \hline
      WESAD \cite{10.1145/3242969.3242985}  & 15 & 13  & \checkmark & \checkmark & \checkmark & \checkmark & \checkmark &  &  & \checkmark & \checkmark &  &  \\
      \hline
      FatigueSet \cite{10.1007/978-3-030-99194-4_14} & 12 & 13  & \checkmark & \checkmark & \checkmark & \checkmark & \checkmark &  & \checkmark & \checkmark &  &  &  \\
      \hline
      MEFAR \cite{DERDIYOK2024109896}      & 23 & 28  & \checkmark & \checkmark & \checkmark & \checkmark & \checkmark &  & \checkmark &  &  &  &  \\
      \hline
      VPFD \cite{10799435}                & 4  & 9   & \checkmark & \checkmark & \checkmark & \checkmark & \checkmark & \checkmark &  &  &  &  &  \\
      \hline
    \end{tabular}
  }
\end{table*}

\section{Heterogeneous and Multi-Source Fatigue Detection Framework}

\subsection{Problem Definition}
We consider a fatigue detection task in the presence of heterogeneous, incomplete, and multi-source sensor data. Specifically, assume there exists:
\begin{itemize}
    \item A target domain dataset  $ \mathcal{D}_T = \{(x_T^{(i)}, y_T^{(i)})\}_{i=1}^{n_T} $  with limited samples, where  $ x_T^{(i)} \in \mathbb{R}^{d_T} $  represents the sensor feature vector defined over  $\mathcal{X}_T$ and  $ y_T^{(i)} \in \mathcal{Y} $  is the label. $n_T$ is the number of target samples and $d_T$ is the number of feature dimensions, i.e., the number of sensors.
     \item A collection of source domains $ \{\mathcal{D}_s\}_{s=1}^{S} $, where each $ \mathcal{D}_s = \{(x_s^{(i)}, y_s^{(i)})\}_{i=1}^{n_s} $ contains samples from a different sensor configuration or environment. Each $ x_s^{(i)} \in \mathbb{R}^{d_s} $ is defined over $\mathcal{X}_s$ and $ y_s^{(i)} \in \mathcal{Y} $. $S$ is the number of source domains, $n_s$ is the number of samples in source domain $s$, and $d_s$ is the feature dimensionality of $\mathcal{D}_s$.
\end{itemize}

In the context of our problem, {heterogeneous} refers to the differences in the feature spaces and data distributions across different domains. Specifically, the feature spaces (i.e., the set of sensors or features) and the data distributions (i.e., the statistical properties of the data) in the target domain and source domains are not identical. This can be formally defined as follows:
\begin{itemize}
   \item {Feature Space Heterogeneity}: For each source $s$, $\mathcal{X}_T \neq \mathcal{X}_s$ may hold while $\mathcal{X}_T \cap \mathcal{X}_s \ne \emptyset$; when $\mathcal{X}_T=\mathcal{X}_s$, our setting reduces to an identical-modality special case (no imputation required).
    \item {Data Distribution Heterogeneity}: $\mathbb{P}_T \neq \mathbb{P}_s$ for $s=1,\ldots,S$, where $\mathbb{P}_T$ and $\mathbb{P}_s$ denote the distributions of the target and source domains, respectively.
\end{itemize}

Since sensors typically differ across domains during data collection, we define

\begin{itemize}

     \item \( \mathcal{X}^\cap := \mathcal{X}_T \cap \mathcal{X}_{s} \)

     \quad (\text{shared modalities between $\mathcal{D}_T$ and  $\mathcal{D}_s$ }) \item \( \mathcal{X}^{s-} := \mathcal{X}_{T} \setminus \mathcal{X}^\cap \)

     \quad (\text{modalities present in $\mathcal{D}_T$ but missing in $\mathcal{D}_s$})
     \item \( \mathcal{X}^{s+} := \mathcal{X}_{s} \setminus \mathcal{X}^\cap \)

     \quad (\text{modalities present in dataset $\mathcal{D}_s$ but  missing in  $\mathcal{D}_T$})

\end{itemize}

Here, we assume that each modality, regardless of its type, is available in at least one domain. Besides, we also have $\mathbb{P}_T^\cap \neq\mathbb{P}_s^\cap $, where $\mathbb{P}_T^\cap$ and $\mathbb{P}_s^\cap$ are the distributions over $\mathcal{X}^\cap$ of $\mathcal{D}_T$ and  $\mathcal{D}_s$, respectively.
 Here, our task is to enhance the performance of a fatigue detector $f$ in the target domain by leveraging data from multiple heterogeneous source domains. Specifically, under distribution shift, we seek to impute the modalities unavailable in the target domain by leveraging multiple heterogeneous source datasets, thereby enhancing the model's fatigue-detection performance. For completeness, for each source $s$ we learn a modality-imputation mapping $g_s:\mathcal{X}^\cap\to\mathcal{X}^{s+}$ on $\mathcal{D}_s$, construct $\hat a_T^{(s)}:=g_s(x^\cap)$, and form the augmented target feature $\tilde x_T^{(s)}=[x^\cap,\hat a_T^{(s)}]$. In the multi-source case, we concatenate or fuse the imputed modalities $\{\hat a_T^{(s)}\}_{s=1}^S$ to obtain the final augmented representation. In this proof-of-concept, the multi-source step is implemented as straightforward sequential enhancement or concatenation, and it does not yet model conflict-aware weighting when different source domains provide inconsistent estimates for the same missing modality.

\subsection{Theoretical Analysis}

To better understand the problem and the proposed solution, we analyze the associated generalization bound. For simplicity, the theoretical analysis considers a single source domain; the multi-source case can be extended in the same spirit. For a random variable $x$ generated from a distribution $\mathbb{P}$, we use $\mathbb{E}_{x\sim \mathbb{P}}$ to denote the expectation over $x$. The expected generalization error of the fatigue detector $f(x)$ is $\mathcal{E}_{\mathbb{P}}\left(f\right)$, while the empirical generalization error is $\mathcal{E}_{\hat{\mathbb{P}}}\left(f\right)$. Throughout this section, $I(x;y)$ denotes the mutual information between the labels $y$ and the training features $x$, i.e., the sensors used in our task.

\begin{lemma}\cite{bu_tightening_2019}
   Suppose the loss function $\mathcal{L}(f(x),y)$ is R-sub-Gaussian under $x\sim\mathbb{P}$ for all $y\in\mathcal{Y}$. Then, we have
\end{lemma}
\begin{equation}
    \mathcal{E}_{\mathbb{P}}\left(f\right)\leq\mathcal{E}_{\hat{\mathbb{P}}}\left(f\right)+\sqrt{\frac{2R^2}{n}\,I(x;y)}
\end{equation}

Let $a$ denote a label-related additional modality and $\hat a$ its imputed counterpart. Define $x^+=[x,a]$ with joint distribution $\mathbb{P}_+$ over $(x^+,y)$, and $\hat x^+=[x,\hat a]$ with joint distribution $\mathbb{Q}$ over $(\hat x^+,y)$.

\begin{theorem}
\label{the1}
    Assume that $x$ and $a$ are conditionally independent given $y$, with $x\sim \mathbb{P}$ and $a\sim\mathbb{Q}_a$, while both are conditionally dependent on $y$. Let the composite observation be defined as $x^+=[x,a]$, where $x^+\sim\mathbb{P}_+$. After incorporating the variable $a$, $\mathcal{E}_{\mathbb{P}_+}$ has a tighter upper bound than $\mathcal{E}_{\mathbb{P}}$. The gap $G$ is
\end{theorem}

\begin{equation}
    G=\Delta  +\sqrt{\frac{2R^2}{n}}\left(\sqrt{I(x^+;y)}-\sqrt{I(x;y)}\right) < 0
\end{equation}

where $\Delta = \mathcal{E}_{\hat{\mathbb{P}}_+}-\mathcal{E}_{\hat{\mathbb{P}}}$.
The proof is given in the Appendix.
Following Theorem \ref{the1}, incorporating label-related features/sensors (i.e., $a$) can reduce the generalization error bound, thereby improving generalization performance. This result supports the use of additional informative modalities in our fatigue-detection task.

\begin{theorem}
\label{the2}
Given a random variable $\hat{a}$ generated from a distribution $\mathbb{Q}_{\hat{a}}$, and $\hat{x}^+=[x,\hat{a}]\sim \mathbb{Q}$, we have
\end{theorem}
\begin{equation}
    \mathcal{E}_{\mathbb{P}_+}\left(f\right)\leq\mathcal{E}_{\hat{\mathbb{Q}}}\left(f\right)+\epsilon_{ideal}+d_{\mathcal{H}\Delta\mathcal{H}}(\mathbb{Q}_{a},\mathbb{Q}_{\hat{a}})+\sqrt{\frac{2R^2}{n}\,I(x;y)}
\end{equation}

where $\mathcal{E}_{\hat{\mathbb{Q}}}\left(f\right)$ is the empirical generalization error over distribution $\mathbb{Q}$, $d_{\mathcal{H}\Delta\mathcal{H}}(\mathbb{Q}_{a},\mathbb{Q}_{\hat{a}})$ is the $\mathcal{H}\Delta\mathcal{H}$ distance between $\mathbb{Q}_{a}$ and $\mathbb{Q}_{\hat{a}}$, and $\mathbb{P}_+$ represents the distribution over the feature $x^+$. The proof is given in the Appendix.

Through Theorems \ref{the1} and \ref{the2}, even without direct access to the additional sensor $a$, we can obtain a tighter generalization error bound if the distribution distance between $\mathbb{Q}_{a}$ and $\mathbb{Q}_{\hat{a}}$ is less than a constant, i.e., $d_{\mathcal{H}\Delta\mathcal{H}}(\mathbb{Q}_{a},\mathbb{Q}_{\hat{a}})\leq \left| \sqrt{\frac{2R^2}{n}}\left(\sqrt{I(x^+;y)}-\sqrt{I(x;y)}\right)-\Delta-\epsilon_{ideal} \right|$. This result suggests that generating such informative modalities can improve fatigue-detection performance.

Thus, to effectively enhance model performance, we are motivated to minimize the distributional discrepancy $d_{\mathcal{H}\Delta\mathcal{H}}(\mathbb{Q}_{a},\mathbb{Q}_{\hat{a}})$ between the imputed modality $\hat{a}$ and the corresponding real modality $a$ by training a regression model $g_s:\mathcal{X}^\cap\xrightarrow{}\mathcal{X}^{s+}$ over $\mathcal{D}_s$. Since the $\mathcal{H}\Delta\mathcal{H}$ distance is symmetric, obeys the triangle inequality, and is bounded, we can regard it as a loss function $\ell$. Following \cite{mansour2009domain}, we obtain the following result.

\begin{theorem}
\label{the3}
 Let $\mathcal{H}$ be a hypothesis class. Let $R_s^\ell(g)$ and $R_T^\ell(g)$ denote the expected loss under $\ell$ for the source and target domains, respectively. Let $g_s^* = \arg\min\limits_{g \in \mathcal{H}} R_s^\ell(g)$ and $g_T^* = \arg\min\limits_{g \in \mathcal{H}} R_T^\ell(g)$ denote the ideal hypotheses for the source and target domains. Then, we have
\end{theorem}

\begin{equation}
 \, R_T^\ell(g_s) \leq R_S^\ell(g_s, g_S^*) + d_{\mathcal{H}\Delta\mathcal{H}}(\mathbb{P}_T^\cap,\mathbb{P}_s^\cap) + \epsilon,
\end{equation}
where $R_S^\ell(g_s, g_S^*) = \mathbb{E}_{x \sim \mathbb{P}_s} \ell(g_s(x), g_s^*(x))$ and $\epsilon = R_T^\ell(g_T^*) + R_S^\ell(g_T^*, g_s^*)$. Since $\hat{a}$ is generated by $g_s$, $d_{\mathcal{H}\Delta\mathcal{H}}(\mathbb{Q}_{a},\mathbb{Q}_{\hat{a}})$ is bounded by $R_S^\ell(g_s, g_S^*)$ and $d_{\mathcal{H}\Delta\mathcal{H}}(\mathbb{P}_T^\cap,\mathbb{P}_s^\cap)$. Following this idea, to reduce the distributional discrepancy $d_{\mathcal{H}\Delta\mathcal{H}}(\mathbb{Q}_{a},\mathbb{Q}_{\hat{a}})$, we also need to reduce the distribution gap $d_{\mathcal{H}\Delta\mathcal{H}}(\mathbb{P}_T^\cap,\mathbb{P}_s^\cap)$.

Taken together, Theorems \ref{the1}--\ref{the3} provide conditional support at the level of the generalization bound for the proposed cross-domain modality-imputation approach. Under the stated assumptions, the target-domain generalization error bound can be tightened when two conditions are met: the added modalities carry fatigue-related information, and discrepancies are reduced between real and imputed modalities and between the source and target domains in the shared feature space. This analysis supports the design choice of learning source-domain modality imputers from shared modalities, but it should not be interpreted as an unconditional guarantee of empirical improvement for every imputation setting.

\subsection{Learning Objective}
 Motivated by Theorems \ref{the1} and \ref{the2}, we present our heterogeneous, multi-source framework. Specifically, given a target domain $\mathcal{D}_T$, our learning objective is to detect fatigue by transferring knowledge from multiple source domains $\{\mathcal{D}_s\}$ for $s = 1, \ldots, S$, where these source domains may contain high-fidelity information extracted from controlled environments. In this formulation, knowledge transfer is implemented by learning from source domains to impute the modalities absent from $\mathcal{D}_T$, i.e., $\mathcal{X}^{s+}$. The learning objective is defined as follows:
\begin{equation}
    \label{eq3}
\begin{aligned}
       \mathcal{L} = \min_{g_s, f} &\underbrace{\sum_{s=1}^S\mathbb{E}_{\mathcal{D}_s} [\mathcal{L}_{rec}(g_s(\mathcal{X}^\cap, \mathcal{X}^{s+})}_{\text{Modality Imputation}} +\underbrace{[\mathcal{L}_{dis}(\mathbb{P}_T^\cap,\mathbb{P}_s^\cap))]}_{\text{Domain Alignment}}\\
       +&\underbrace{ \mathbb{E}_{\mathcal{D}_T}[\mathcal{L}_{task}(f(\tilde{\mathcal{X}_T}), y)]}_{\text{Task Aware Detection}}
\end{aligned}
\end{equation}

where $g_s$ is the modality imputer used to estimate $\mathcal{X}^{s+}$, and $f$ is the classifier for fatigue detection. $\mathcal{L}_{rec}$ represents the imputation loss (e.g., MSE for regression). $\mathcal{L}_{task}$ denotes the task-specific loss (e.g., MSE for regression and CE for classification), and $\tilde{\mathcal{X}_T}$ represents the imputation-enhanced data from domain $\mathcal{D}_T$. This data is enhanced or completed using the information estimated by the modality imputer $g_s$ from other domains.

Moreover, motivated by Theorem \ref{the3}, we introduce a domain-alignment term to reduce $d_{\mathcal{H}\Delta\mathcal{H}}(\mathbb{P}_T^\cap,\mathbb{P}_s^\cap)$. Specifically, a batch-normalization step, referred to as \textit{BatchNorm}, is applied to both the implemented imputer and downstream classifier to stabilize intermediate feature representations \cite{ioffe2015batchnormalizationacceleratingdeep,li2016revisitingbatchnormalization}. Additionally, to optimize the objective function in Equation \ref{eq3}, particularly the task-specific loss, we adopt the Jacobian norm regularization method \cite{Li_2023}, referred to later as \textit{JacobianNorm}, and apply it to all downstream classifiers evaluated in the classification stage.


Note that this objective is optimized in separate stages rather than through end-to-end joint training. For each auxiliary source domain, the imputer $g_s$ is first trained on the modalities shared with the target domain as inputs and the modality available only in that source domain as the imputation target. The trained imputer is then applied to the target-domain shared modalities to impute the unavailable modality, after which the fatigue classifier $f$ is trained on the final enhanced target-domain representation.

\subsection{Framework}
\begin{figure*}[!t]
    \centering
    \includegraphics[width=\textwidth]{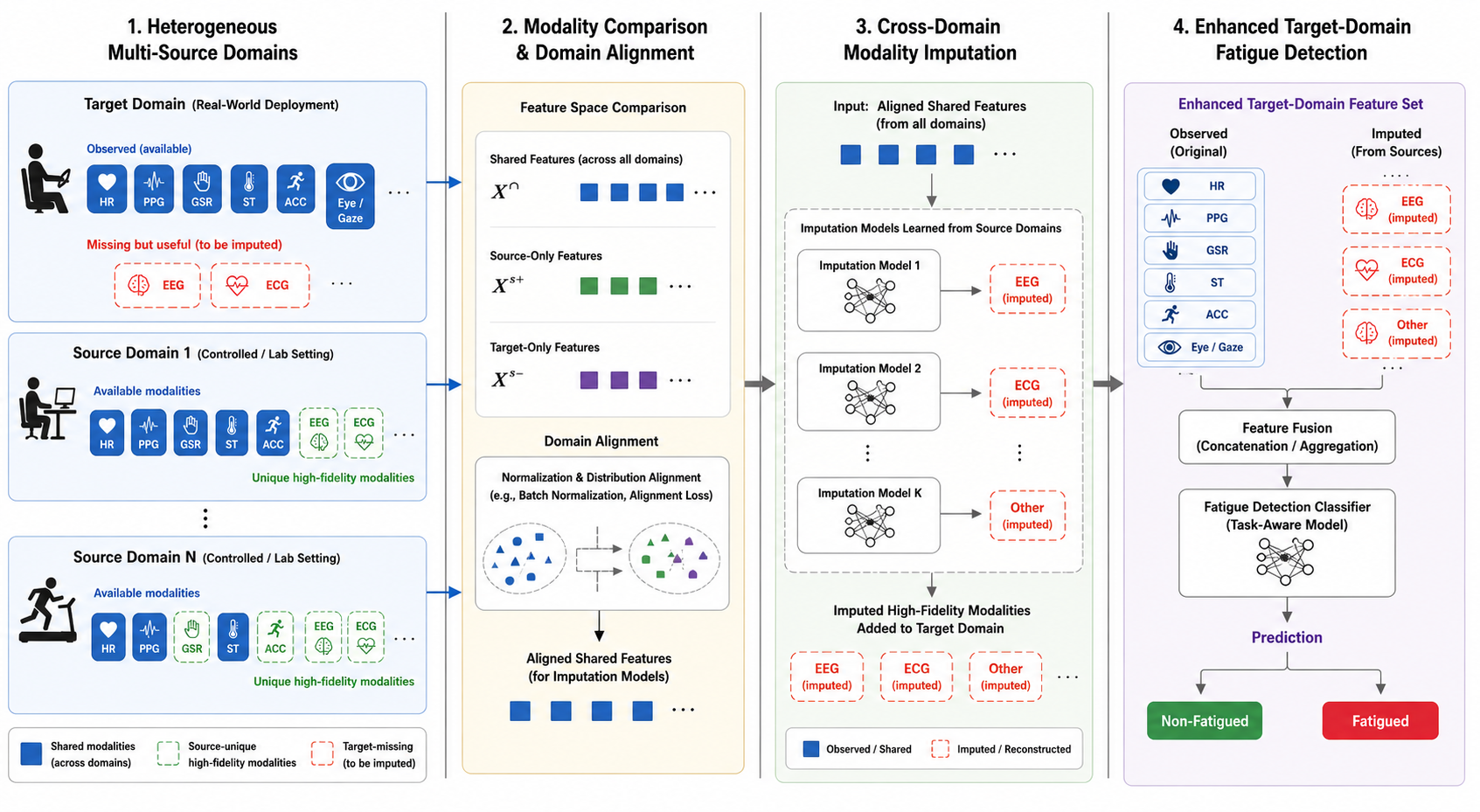}
    \caption{High-level overview of the proposed framework. Blue solid boxes denote available modalities observed in source or target domains, and orange dashed boxes denote imputed modalities transferred to the target domain. In the reported experiments, VPFD serves as the target domain, MEFAR provides EEG information for imputation, and FatigueSet provides ECG information for imputation. The framework first leverages high-fidelity source domains, then performs cross-domain modality imputation, fuses the available and imputed target-domain features, and finally makes the target-domain fatigue decision.}
    \label{fig:method_overview}
\end{figure*}

\begin{figure}[!t]
    \centering
    \includegraphics[width=0.65\linewidth]{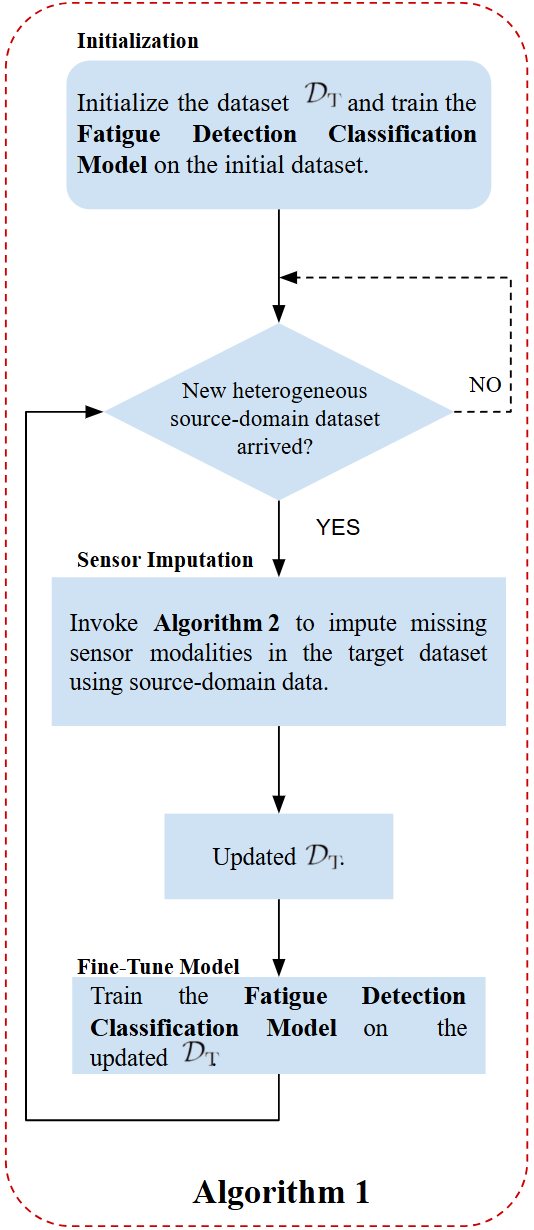}
    \caption{Sequential multi-source enhancement workflow. In the reported experiments, VPFD is the target domain, and its representation is enhanced by appending imputed EEG from MEFAR and imputed ECG from FatigueSet to the available target-domain modalities.}
    \label{fig:overall_framework}
\end{figure}

\begin{figure}[!t]
    \centering
    \includegraphics[width=\linewidth]{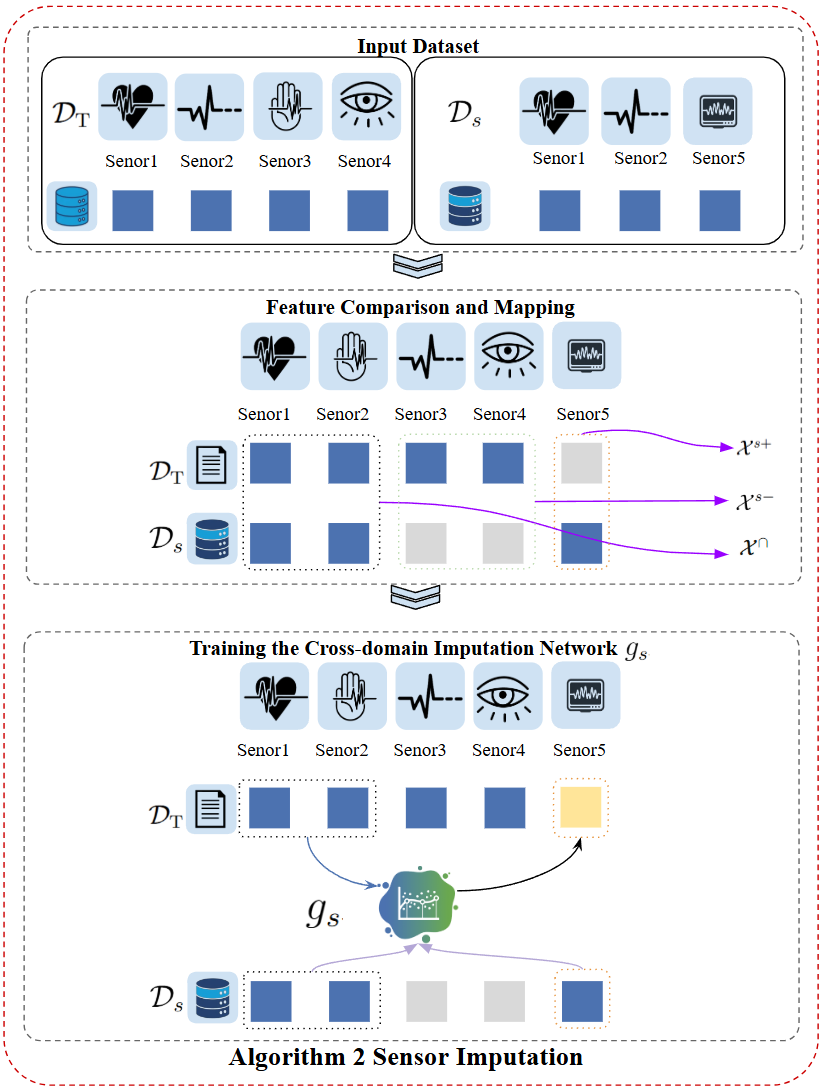}
    \caption{Modality imputation module. Shared modalities such as HR, PPG, GSR, ST, and ACC are used as inputs for imputation, while EEG from MEFAR and ECG from FatigueSet are learned as imputation targets for VPFD.}
    \label{fig:Cross}
\end{figure}

\begin{figure}[!t]
    \centering
    \includegraphics[width=\linewidth]{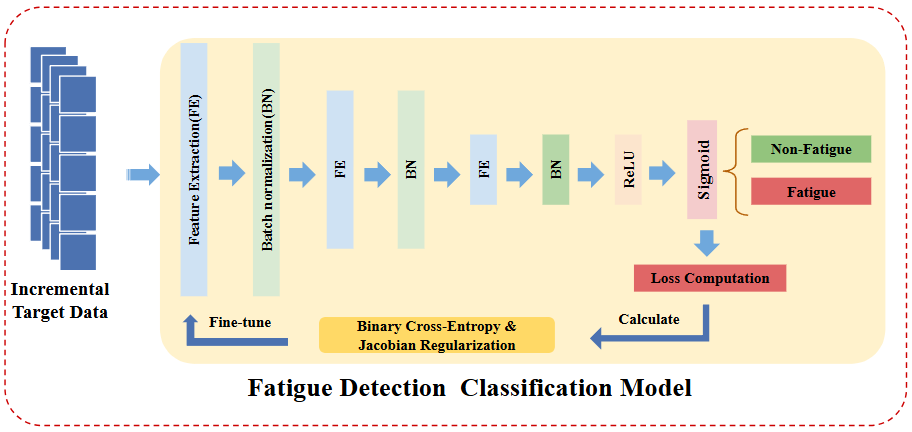}
    \caption{Fatigue detection model. For VPFD, the classifier uses both available target-domain modalities and imputed EEG and/or ECG modalities to predict the fatigue state.}
    \label{fig:fatigue}
\end{figure}

For the problem defined above, we propose a framework that enables cross-domain modality imputation to improve fatigue detection under heterogeneous sensor configurations. A high-level overview is provided in Fig.\ref{fig:method_overview}. The sequential multi-source enhancement workflow is then detailed in Fig.\ref{fig:overall_framework} and Algorithm 1.
The method is designed to handle data collected from different sensor configurations with partially overlapping feature spaces. When new data from a different sensor setup become available, the framework compares its feature space with that of the target domain. If discrepancies are found, i.e., missing modalities, the method initiates a targeted imputation process to infer the missing modalities. This imputation process is detailed in Algorithm 2 and illustrated in Fig.\ref{fig:Cross}. The aligned and imputation-enhanced data are then incorporated into the model to enhance prediction accuracy in the target domain, as illustrated in Fig.\ref{fig:fatigue}.

\begin{algorithm}[H]
\caption{Multi-Source Fatigue Detection}
\begin{algorithmic}[1]
\label{alg:enhance_target}
\REQUIRE Target dataset $\mathcal{D}_T$, source datasets $\{\mathcal{D}_s\}_{s=1}^{S}$
\ENSURE Detector $f$ trained on enhanced target data

\FOR{$s = 1$ {\bf to} $S$}
    \STATE Compute shared and unshared features: $\mathcal X^{s-}, \mathcal X^{s+}$ w.r.t.\ $(\mathcal{D}_T, \mathcal{D}_s)$
    \STATE $\mathcal{D}_T \gets \textsc{ModalityImpute}(\mathcal{D}_T, \mathcal{D}_s)$
\ENDFOR
\STATE Train detector $f_\phi$ on final enhanced $\mathcal{D}_T$
\end{algorithmic}
\end{algorithm}

\begin{algorithm}[H]
\caption{\textsc{ModalityImpute}$(\mathcal{D}_{\text{T}}, \mathcal{D}_s)$}
\label{alg:sensor_imputation}
\begin{algorithmic}[1]
\REQUIRE Reference dataset $\mathcal{D}_{\text{T}}$, auxiliary source dataset $\mathcal{D}_s$
\ENSURE  Updated target dataset $\mathcal{D}_{\text{T}}$

\STATE Compute modality difference sets
      \[
         \mathcal{X}^{s-}= \mathcal{X}_{\text{T}}\setminus\mathcal{X}^\cap, \qquad
          \mathcal{X}^{s+} := \mathcal{X}_{s} \setminus \mathcal{X}^\cap
      \]

\IF{$\mathcal{X}^{s+}\neq\emptyset$}

    \STATE Apply $g_s$ to impute the missing modalities $\mathcal{X}^{s+}$ in $\mathcal{D}_T$ based on $\mathcal{D}_s$, and update $\mathcal{D}_T$.

\ENDIF

\STATE {return} $\mathcal{D}_{\text{T}}$
\end{algorithmic}
\end{algorithm}

\section{Experiments}

A series of experiments are designed to validate our framework. Section IV.A details the experimental setup, including the datasets, environment, and data preprocessing. Section IV.B describes the experimental design and discusses the results.

\subsection{Experimental Setup}\label{ES}

For the experiments described in the next subsection, three datasets are selected because they contain diverse physiological signals, with some modalities overlapping across datasets and others being unique. Although these datasets may contain the same signal types, they often originate from different sensing platforms, resulting in differences in sampling rates, noise patterns, and numerical scales. Therefore, a unified preprocessing procedure is needed to standardize the signals before experimentation. After preprocessing, multiple widely used classification methods are trained on both individual datasets and the enhanced target data. To that end, this subsection describes the selected datasets, preprocessing steps, and implementation environment in detail.

\subsubsection{Datasets}\label{ES1}
Out of all publicly available datasets for
fatigue, VPFD\cite{10799435}, MEFAR\cite{DERDIYOK2024109896}, and FatigueSet\cite{kalanadhabhatta2021fatigueset} were chosen for our experiments because they share a large set of overlapping physiological signals while each also contains unique ones. For instance, all three datasets include HR, PPG, GSR, ST, and ACC. Additionally, MEFAR provides EEG
data, FatigueSet includes both EEG and ECG data, and VPFD offers unique eye-tracking data.

\begin{itemize}
\item
The VPFD dataset\cite{10799435} is a multimodal fatigue-detection dataset designed to reflect practical deployment conditions. It was collected using wearable devices such as the HoloLens and Google Pixel Watch 2. The dataset contains physiological signals from four 27-year-old participants monitored for up to 9 hours during sports, driving, and research activities.

\item
The MEFAR dataset\cite{DERDIYOK2024109896} focuses on occupational mental-fatigue analysis. It comprises neurophysiological data from 23 participants across four occupational groups, i.e., academicians, technicians, engineers, and kitchen workers. The EEG signals were captured by the NeuroSky MindWave, while other signals were recorded by the Empatica E4 wristband.

\item
The FatigueSet dataset\cite{kalanadhabhatta2021fatigueset} contains data from 12 participants performing controlled physical and cognitive tasks. It utilizes the Muse S headband for EEG signals, the Zephyr BioHarness chestband for ECG signals, and the Empatica E4 wristband for the remaining physiological signals.
\end{itemize}

\subsubsection{Data Preprocessing} \label{ES3}
To minimize noise and artifacts, a series of tailored preprocessing techniques is first applied to each signal type. ACC and PPG signals are processed using recursive least squares (RLS) filtering and singular spectrum analysis (SSA) to remove motion artifacts while preserving critical physiological features. HR, GSR, and ST signals undergo maximum-outlier filtering to eliminate extreme values and ensure data smoothness. Following noise reduction, all signals are then resampled to 32 Hz to standardize the sampling rate across devices. 

After resampling, signals are aligned across datasets by physiological category. Scaling is fitted separately within each dataset: min-max scaling parameters are estimated once for each signal category and then applied uniformly to all values of that signal category within the dataset. The target scaling range is kept consistent across datasets for the same physiological category: HR and ST are scaled to $[0,1]$, while EEG, ECG, GSR, ACC, EYE, and PPG are scaled to $[-1,1]$. These scaling parameters are not estimated separately for individual participants, training portions, or test portions.

Finally, a block-based temporal split is used to divide each preprocessed dataset into an 80\% training set and a 20\% test set. Here, a block is defined as a temporally continuous segment from the same participant under the same fatigue-label condition. For each block, the first 10\% and last 10\% are assigned to the test set, while the middle 80\% is assigned to the training set. This preserves temporal order and avoids random shuffling. Windowing is then performed after this block-level split. For the training portion, 10-s windows are extracted from the middle 80\% using a 3-s sliding step, meaning that consecutive training windows overlap by 7 s. This overlap is intentionally introduced to preserve temporal continuity between adjacent training samples, allowing the model to learn fatigue-related temporal relationships rather than treating each window as an isolated segment. For the test portions, 10-s windows are extracted from the first and last 10\% using a non-overlapping scheme, i.e., each test window advances by the full 10 s. Each 10-s window is represented as a resampled signal sequence, rather than being reduced to window-level statistical features. This ensures that training and test windows are generated from temporally separate portions of each block. After window extraction, 10\% of the training windows within each labeled block are reserved for validation.

\subsubsection{Implementation Environment} \label{ES2}

Experiments are executed on a PC equipped with an Intel Core i9-14900K processor, an NVIDIA RTX 4090 GPU, and 128 GB of RAM. The system operates on Windows 11, with a software stack comprising Python 3.10.15, PyTorch 2.4.1, scikit-learn 1.1.3, pandas 2.2.2, and NumPy 1.26.4.

The experiments include two learning stages. The imputation stage predicts continuous physiological signals for masked or unavailable modalities, while the classification stage predicts binary fatigue labels from the original or imputation-enhanced feature sequences. Tables II and III summarize the main model architecture settings and training hyperparameters used across these two stages, respectively. A fixed random seed of 42 is used for model initialization and stochastic training operations. For JacobianNorm regularization in the classification stage, the weight is set to 0.008.

\begin{table}[!t]
\centering
\caption{Main Model Architecture Settings.}
\label{tab:model_settings}
\footnotesize
\setlength{\tabcolsep}{4pt}
\renewcommand{\arraystretch}{1.15}
\begin{tabularx}{\columnwidth}{@{}l l X@{}}
\toprule
\textbf{Scope} & \textbf{Model} & \textbf{Main Settings} \\
\midrule
\multicolumn{3}{@{}l}{\textit{Shared by imputation and classification}} \\
Both & MLP &
Two hidden layers with 128 and 64 units \\
Both & LSTM &
Hidden size 128 with a 64-unit task-specific head \\
Both & CNN1D &
Three temporal convolution layers with 64, 128, and 256 channels; kernels 5, 3, and 3 \\
\midrule
\multicolumn{3}{@{}l}{\textit{Stage-specific architectures}} \\
Imputation & TimeMixer &
Hidden dimension 128; feed-forward dimension 256; 8 heads; 2 encoder layers; dropout 0.1; kernels 5 and 3 \\
Classification & Transformer &
Hidden dimension 128; 8 heads; 3 encoder layers; feed-forward dimension 512 \\
\bottomrule
\end{tabularx}
\end{table}

\begin{table}[!t]
\centering
\caption{Hyperparameters for Imputation and Classification.}
\resizebox{\columnwidth}{!}{%
\begin{tabular}{lcc}
\hline
\textbf{Parameter} & \textbf{Imputation Phase} & \textbf{Classification Phase} \\ \hline
Learning rate & 1e-3 & 3e-4 \\
Batch size & 512 & 256 \\
Optimizer & Adam & Adam \\
Dropout rate & - & 0.2 \\
Max epochs & 2000 & 200 \\
Early stopping patience & 40 & 20 \\ \hline
\end{tabular}
}
\label{tab:chicp}
\end{table}

\subsection{Experimental Design and Results}\label{EDR}

\subsubsection{Regression Based Imputation Model Selection}
Without loss of generality, the first set of experiments validates our imputer selection by assessing its ability to recover a single masked modality. We compare four regression architectures--MLP, LSTM (Long Short-Term Memory), CNN1D, and TimeMixer\cite{wang2024timemixerdecomposablemultiscalemixing}--on two datasets: FatigueSet (with ECG masked) and MEFAR (with EEG masked). In each case, the missing
modality is replaced by the model prediction, and we record the mean squared error (MSE). All errors are computed on strictly held-out, non-overlapping 10-s windows sampled at 32 Hz. The results show that the MLP consistently yields the lowest MSE, as shown in Table~\ref{tab:test_performance}. Consequently, the MLP serves as the default model for all subsequent modality-imputation tasks unless otherwise specified.

\begin{table}[htbp]
    \centering
    \caption{Regression performance of different models}
    \label{tab:test_performance}
    \small
    \setlength{\tabcolsep}{4pt}
    \begin{tabular}{lcc}
        \hline
        {Model} & {FatigueSet-ECG(MSE)} & {MEFAR-EEG(MSE)} \\
        \hline
        CNN1D          & 0.5063 & 0.1031 \\
        MLP            & 0.4074 & 0.1028 \\
        LSTM           & 0.4250 & 0.1076 \\
       TimeMixer           & 0.5508 & 0.5455 \\
        \hline
    \end{tabular}
\end{table}

\subsubsection{Validity of Regression‑Imputed Data Versus Noise Baselines}
The second set of experiments is designed not only to verify Theorems 1 and 2 but also to demonstrate that imputed data can effectively support classification tasks. In this case, VPFD is chosen with its gaze modality masked and imputed using three different methods: MLP-predicted values, MLP prediction with Gaussian noise, and pure random noise. We refer to the modified datasets as $\text{VPFD}_{\text{MLP}}$, $\text{VPFD}_{\text{MLP+noise}}$, and $\text{VPFD}_{\text{noise}}$, respectively. The added noise follows a zero-mean Gaussian distribution whose standard deviation is set to twice the maximum absolute value of the original VPFD gaze signal. Table~\ref{tab:results2} shows that the classifier trained on regression-generated data achieves performance comparable to training on the original data, highlighting the utility of imputation for missing modalities. By contrast, injecting noise--partially or entirely--increases the distributional mismatch and degrades classifier performance.

\begin{table}[h]
    \centering
    \caption{Classification performance on different training datasets.}
    \begin{tabular}{lcc}
        \hline
        {Training Dataset} & {Cross-Entropy Loss} & {Accuracy (\%)} \\
        \hline
        $\text{VPFD}$ & 1.2789 & 81.15 \\
        $\text{VPFD}_{\text{MLP}}$ & 1.5523 & 80.00 \\
        $\text{VPFD}_{\text{MLP+noise}}$ & 1.2041 & 60.58 \\
        $\text{VPFD}_{\text{noise}}$ & 1.6507 & 60.15 \\
        \hline
    \end{tabular}

    \label{tab:results2}
\end{table}

\subsubsection{Impact of Cross-Domain Imputed Modalities on Classification Performance}
The third set of experiments evaluates the \textsc{VPFD} dataset across different scenarios in which modalities are imputed from two auxiliary sources, FatigueSet and MEFAR, using different auxiliary-source and imputer configurations, and then evaluated with multiple classifiers.

We first train the fatigue detector \(f_\phi\) on the unmodified VPFD to obtain the baseline. Next, we run Algorithm~\ref{alg:enhance_target} under three configurations of auxiliary-source availability, with the MLP and TimeMixer serving as imputers, respectively: (a) only FatigueSet is available, which is used to impute the missing ECG modality, resulting in the variants VPFD w/ ECG (MLP) and VPFD w/ ECG (TimeMixer); (b) only \textsc{MEFAR} is available, which is used to impute the missing EEG modality, resulting in the variants VPFD w/ EEG (MLP) and VPFD w/ EEG (TimeMixer); and (c) both FatigueSet and \textsc{MEFAR} are available, enabling the sequential imputation of ECG and EEG for the VPFD w/ ECG+EEG (MLP) and VPFD w/ ECG+EEG (TimeMixer) variants. In all cases, the original VPFD features are preserved, and only the missing modalities are imputed and appended. Again, each dataset is split, with 80\% used to train a classifier and the remaining 20\% used for evaluation.

To assess the behavior of our approach across model architectures, we train four different classifiers in this experiment--MLP, LSTM, CNN1D, and a Transformer\cite{vaswani2017attention}--on each dataset variant. We assess performance using multiple metrics, including accuracy, binary F1, area under the ROC curve (ROC-AUC), area under the precision-recall curve (PR-AUC), macro F1, weighted F1, expected calibration error (ECE), and the Brier score. ECE is computed with 15 uniform confidence bins. This setup allows us to evaluate not only how each additional imputed modality, individually or in combination, contributes to downstream fatigue-detection performance, but also how the framework performs across diverse classifier types.

Table~\ref{tablepwwww} summarizes the classification metrics under these scenarios. Two main observations can be drawn from the results. First, the combined EEG+ECG setting yields more pronounced improvements in several classifier settings, especially relative to the EEG-only imputation setting, although the gains are not uniform across all imputers and classifiers. This pattern suggests that imputed EEG and ECG can provide complementary information for VPFD classification, while the benefit depends on the imputer and classifier used. Second, performance depends on the imputer-classifier pairing: using TimeMixer as the imputer or Transformer as the classifier does not by itself consistently improve performance, whereas combining the TimeMixer imputer with the Transformer classifier gives strong results in several imputation-enhanced VPFD settings. Overall, these results suggest that imputed modalities can improve VPFD classification, but the magnitude of the gain depends on both the source modality and the downstream classifier.

\newcommand{\metricscol}{\scriptsize\makecell[c]{
Acc\\ F1$_{\text{bin}}$\\ ROC-AUC\\ PR-AUC\\
F1$_{\text{macro}}$\\ F1$_{\text{weighted}}$\\
ECE\\ Brier}}

\newcommand{\mvals}[8]{\scriptsize\makecell[c]{
#1\\ #2\\ #3\\ #4\\ #5\\ #6\\ #7\\ #8}}

\begin{table}[htbp]
    \centering
    \caption{Classification metrics (\%) under VPFD imputed-modality scenarios.}
    \label{tablepwwww}
    \scriptsize
    \setlength{\tabcolsep}{3.5pt}
    \renewcommand\arraystretch{1.15}
    \resizebox{\linewidth}{!}{
    \begin{tabular}{cccccc}
        \hline
        {Dataset} & {Metrics} & {MLP} & {LSTM} & {CNN1D} & {Transformer} \\
        \hline

        VPFD &
        \metricscol &
        \mvals{81.47}{80.10}{88.20}{93.57}{86.50}{86.50}{5.77}{9.63} &
        \mvals{83.61}{82.90}{90.10}{93.71}{89.18}{89.18}{6.86}{8.26} &
        \mvals{60.34}{59.20}{64.80}{91.42}{90.53}{90.53}{5.08}{7.53} &
        \mvals{82.31}{81.72}{91.84}{91.81}{82.30}{82.29}{14.08}{15.31} \\
        \hline

        VPFD w/ EEG (MLP)&
        \metricscol &
        \mvals{84.13}{83.30}{90.70}{82.51}{76.59}{76.59}{7.32}{16.35} &
        \mvals{88.77}{88.10}{94.00}{83.32}{77.14}{77.14}{7.97}{15.66} &
        \mvals{81.75}{80.90}{89.10}{84.16}{76.22}{76.22}{7.51}{16.22} &
        \mvals{84.98}{84.98}{89.81}{81.63}{78.19}{78.20}{19.10}{19.88} \\
        \hline
        VPFD w/ EEG (TimeMixer) &
        \metricscol &
        \mvals{82.90}{84.30}{86.60}{82.51}{76.59}{76.59}{7.32}{16.35} &
        \mvals{81.00}{81.90}{83.90}{83.32}{77.14}{77.14}{7.97}{15.66} &
        \mvals{73.50}{78.20}{79.80}{84.16}{76.22}{76.22}{7.51}{16.22} &
        \mvals{89.17}{90.24}{96.17}{95.47}{89.04}{89.10}{9.32}{9.23} \\
        \hline

        VPFD w/ ECG (MLP)&
        \metricscol &
        \mvals{84.14}{83.40}{90.60}{86.90}{75.51}{75.70}{7.78}{14.97} &
        \mvals{86.99}{86.20}{93.20}{88.36}{80.67}{80.78}{7.76}{13.66} &
        \mvals{81.55}{80.60}{88.80}{88.41}{79.81}{79.92}{6.74}{13.88} &
        \mvals{81.75}{82.93}{88.16}{95.01}{88.88}{88.92}{9.76}{10.15} \\
        \hline
        VPFD w/ ECG (TimeMixer) &
        \metricscol &
        \mvals{85.40}{86.73}{93.96}{91.83}{85.25}{85.25}{5.18}{9.80} &
        \mvals{88.98}{90.05}{94.39}{91.88}{88.85}{88.86}{5.84}{8.65} &
        \mvals{90.84}{91.06}{95.68}{93.94}{90.84}{90.84}{3.93}{7.01} &
        \mvals{92.56}{92.54}{97.45}{97.30}{92.56}{92.56}{6.32}{6.78} \\
        \hline

        VPFD w/ EEG+ECG (MLP) &
        \metricscol &
        \mvals{91.13}{90.80}{96.20}{89.47}{81.39}{81.45}{10.48}{14.04} &
        \mvals{92.12}{91.60}{96.90}{86.97}{76.67}{76.75}{8.46}{15.82} &
        \mvals{88.43}{87.90}{94.40}{86.39}{79.58}{79.67}{11.01}{15.41} &
        \mvals{85.03}{86.34}{92.20}{91.83}{84.90}{84.95}{13.01}{13.86} \\
        \hline

        VPFD w/ EEG+ECG (TimeMixer) &
        \metricscol &
        \mvals{87.80}{88.69}{95.39}{93.66}{87.73}{87.73}{5.39}{8.37} &
        \mvals{92.61}{92.80}{96.88}{96.30}{92.60}{92.60}{5.64}{6.14} &
        \mvals{92.29}{92.57}{95.08}{88.44}{92.28}{92.28}{4.11}{6.49} &
        \mvals{92.86}{93.04}{97.88}{97.71}{92.85}{92.85}{6.38}{6.50} \\
        \hline
    \end{tabular}
    }
\end{table}

\subsubsection{Ablation of Batch Normalization and Jacobian Regularization}
The fourth experiment, summarized in Table \ref{tab:ablation_accuracy}, conducts a classifer-side ablation study to examines how incorporating  BatchNorm and JacobianNorm into the downstream classifiers influence classification performance. The MLP classifier is used as the controlled testbed in this experiment. The ablation study compares four settings: Batch Normalization (BN) alone, Jacobian-norm regularization alone, both combined, and neither (serving as the baseline). BN stabilizes the classifier representation, while the Jacobian term regularizes the classifier's sensitivity to input perturbations. When applied together, these mechanisms provide a stable representation space and smoother input-output sensitivity, leading to the best overall accuracy. This complementary effect is particularly evident in the configuration where both auxiliary sources are available, as the inclusion of multiple imputed modalities (i.e., EEG+ECG) accentuates the performance gap between models with and without these regularization components.

\begin{table}[h]
    \centering
    \caption{Ablation study results (Accuracy \%)}
    \label{tab:ablation_accuracy}
    \resizebox{\columnwidth}{!}{
    \begin{tabular}{lccc}
        \hline
        {Method} & {VPFD w/ EEG} & {VPFD w/ ECG} & {VPFD w/ EEG+ECG} \\
        \hline
        Baseline              & 81.78 & 81.20 & 83.68 \\
        BN                    & 82.42 & 81.40 & 85.83 \\
        Jacobian              & 83.28 & 83.60 & 87.35 \\
        BN + Jacobian         & 84.14 & 84.13 & 91.13 \\
        \hline
    \end{tabular}
    }
\end{table}

\section{Conclusion}

Fatigue detection is important in high-stakes environments, where reduced alertness can have serious consequences. It is also an important human-state modeling problem for human-machine systems, where timely fatigue estimation can support warning, scheduling, intervention, and safety-management decisions. Despite advances in sensor technology, reliability and accessibility in real-world environments remain major challenges, as many high-end sensors, such as EEG, are expensive, highly sensitive to environmental conditions, and require substantial operational setup outside controlled settings. To address this issue, this paper proposes a proof-of-concept framework that enables knowledge transfer across differently instrumented sensor domains, aiming to improve fatigue detection in real-world sensor-constrained scenarios. In particular, our approach
transfers knowledge learned from high-fidelity laboratory sensors
to field-deployable systems by using shared modalities
as a bridge, thereby maintaining detection accuracy when
some modalities are unavailable.
Our experiments provide empirical support for this claim and show that transferred knowledge can help improve detection when high-end sensors are unavailable. 

The research can be extended in several directions. For instance, as a proof of concept, VPFD is treated in our experiments as a field-oriented dataset with a limited number of participants. Thus broader validation with larger cohorts remains an important direction for future work. Although this paper focuses on the framework, our experiments
show that incorporating more advanced algorithms can
offer additional benefits. In particular, we observe that using
TimeMixer and a Transformer yields small but consistent gains
over traditional configurations. At the same time, the current proof-of-concept handles multiple source domains through simple sequential enhancement and straightforward fusion; it does not yet model conflict-aware weighting when different sources provide inconsistent estimates for the same missing modality. We plan to explore this direction in future work by developing jointly optimized imputation and classification architectures with adaptive multi-source weighting to better capture nonlinear dependencies and improve fatigue-detection performance. We also aim to further examine its real-time deployment aspects, focusing on computational efficiency, latency, and resource requirements to ensure suitability for field applications.

\bibliographystyle{elsarticle-num}

\bibliography{ref}
\begin{IEEEbiography}
[{\includegraphics[width=1in,height=1.25in,clip,keepaspectratio]{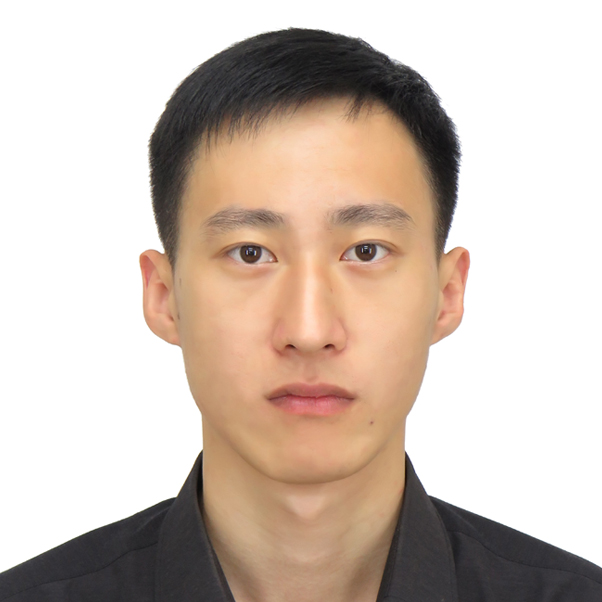}}]
{LuoBin Cui} (Member, IEEE) received the B.S. degree from Jiangsu University of Technology, P.R. China, in 2020, and the M.S. degree from Monmouth University, USA, in 2022. He is currently pursuing the Ph.D. degree in Electrical and Computer Engineering at Rowan University, USA. His research interests include machine learning and artificial intelligence, with a particular focus on graph neural networks and their applications. He has also conducted research on fatigue detection using physiological data and explored the development of educational games to enhance student comprehension in science, technology, and engineering fields.
\end{IEEEbiography}
\begin{IEEEbiography}
[{\includegraphics[width=1in,height=1.25in,clip,keepaspectratio]{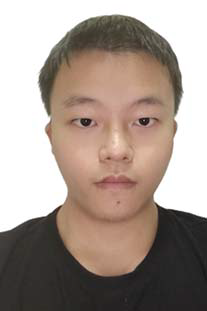}}]
{Yanlai Wu} received the B.S. degree in Artificial Intelligence from the School of Information Science and Engineering, Chongqing Jiaotong University, Chongqing, China. He is currently pursuing the M.S. degree in Electrical and Computer Engineering with the School of Engineering, Rowan University, USA.

His current research interests include pattern recognition, computer vision, and natural language processing.
\end{IEEEbiography}

\begin{IEEEbiography}
[{\includegraphics[width=1in,height=1.25in,clip,keepaspectratio]{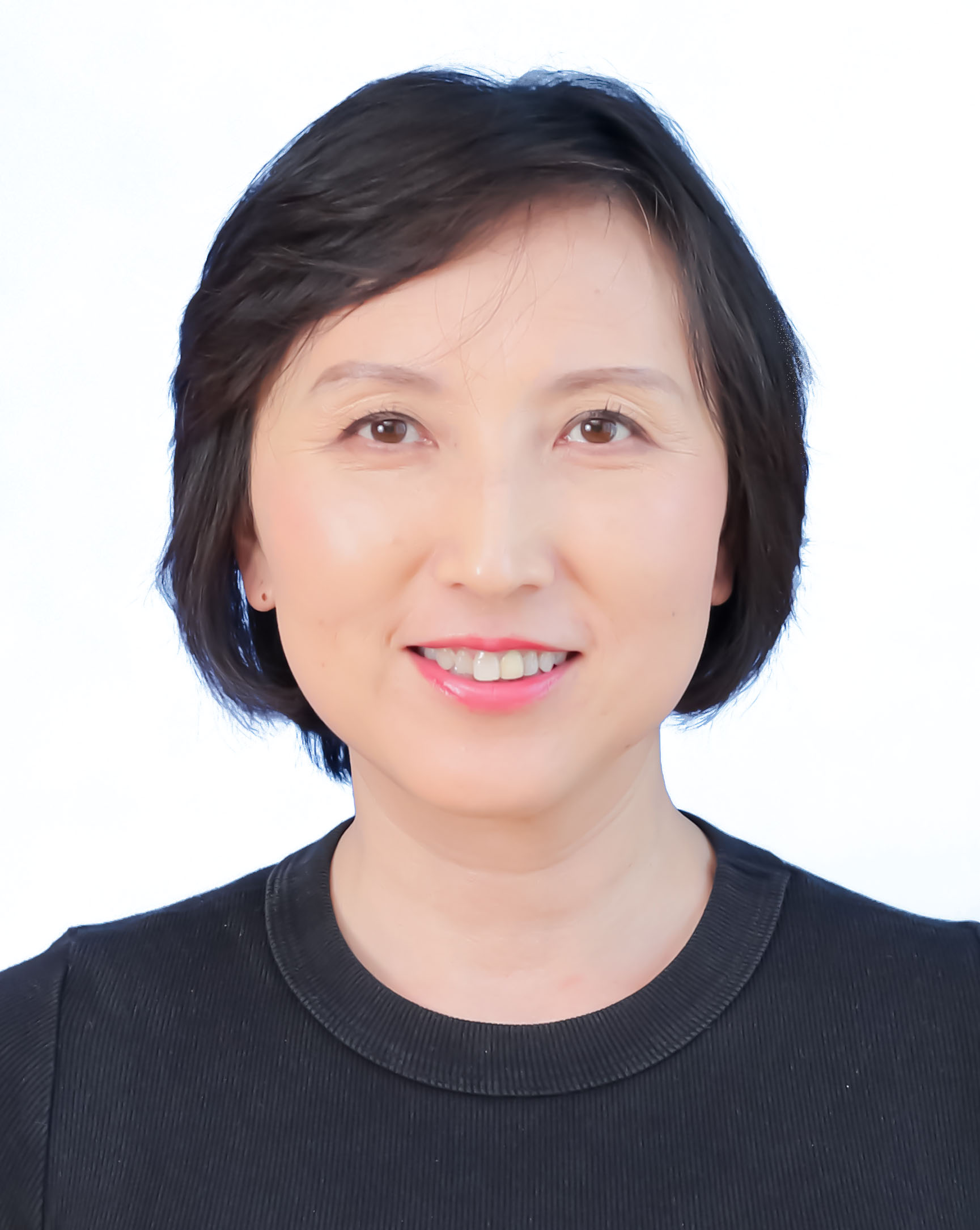}}]
{Ying Tang} (Senior Member, IEEE) received the B.S. and M.S. degrees from Northeastern University, P. R. China, in 1996 and 1998, respectively, and the Ph.D. degree from the New Jersey Institute of Technology in 2001. She is currently a Full Professor and the Undergraduate Program Chair of Electrical and Computer Engineering at Rowan University, Glassboro, New Jersey. Her current research interests lie in the areas of cyber-physical social systems, extended reality, adaptive and personalized systems, modeling and adaptive control for computer-integrated systems, and sustainable production automation. Her work has been continuously supported by NSF, EPA, the U.S. Army, FAA, DOT, private foundations, and industry. She holds three U.S. patents and has over 250 peer-reviewed publications, including 88 journal articles, 2 edited books, and 6 book/encyclopedia chapters. Dr. Tang is presently an Associate Editor of IEEE Transactions on Systems, Man, and Cybernetics: Systems, IEEE Transactions on Intelligent Vehicles, IEEE Transactions on Computational Social Systems, and Springer's Discover Artificial Intelligence. She is the Founding Chair of the Technical Committee on Intelligent Solutions to Human-aware Sustainability for IEEE Systems, Man, \& Cybernetics Society and the Founding Chair of the Technical Committee on Sustainable Production Automation for the IEEE Robotics and Automation Society.
\end{IEEEbiography}
\begin{IEEEbiography}
[{\includegraphics[width=1in,height=1.25in,clip,keepaspectratio]{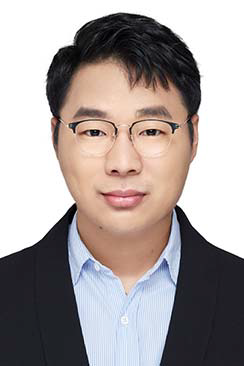}}]
{Weikai Li} received his B.S. degree in Information and Computing Science (2015) and M.S. degree in Computer Science and Technology (2018) from Chongqing Jiaotong University, Chongqing, China. He obtained his Ph.D. degree in Computer Science and Technology from the College of Computer Science and Technology, Nanjing University of Aeronautics and Astronautics, Nanjing, China, in 2022.

He is currently a Lecturer at the School of Computer and Artificial Intelligence, Shandong Jianzhu University. He is also a Researcher with the MIIT Key Laboratory of Pattern Analysis and Machine Intelligence (PAMI), Nanjing, and China Science IntelliCloud Technology, Hefei, China.

His research interests include pattern recognition and machine learning.

\end{IEEEbiography}

\newpage
\appendix

\section{Theoretical Analysis}

\renewcommand{\thelemma}{\arabic{lemma}}
\setcounter{lemma}{0}
\begin{lemma}\cite{bu_tightening_2019}
\label{lemm2}
   Suppose the loss function $\mathcal{L}(f(x),y)$ is R-sub-Gaussian under $x\sim\mathbb{P}$ for all $y\in\mathcal{Y}$, then, we have:
\end{lemma}
\begin{equation}
    \mathcal{E}_{\mathbb{P}}\left(f\right)\leq\mathcal{E}_{\hat{\mathbb{P}}}\left(f\right)+\sqrt{\frac{2R^2}{n} I(x,y)}
\end{equation}
where $n$ is the number of the training samples and $I(x,y)$ is the mutual information between $x$ and $y$.
\renewcommand{\thetheorem}{\arabic{theorem}}
\setcounter{theorem}{0}

\begin{theorem}
\label{app:the1}
   Assume that  $x,a$ are conditionally independent given  $y$,$x\sim \mathbb{P},a\sim\mathbb{Q}_a$,  while both being conditionally dependent on variable of $y$ ,  Let the composite observation be defined as $x^+=[x,s],x^+\sim\mathbb{P}_+$, with the incorporating the variable $a$ , the $\mathcal{E}_{\mathbb{P}_+}$ have a tighten upper bound than $\mathcal{E}_{\mathbb{P}}$, the gap $G$ is:
\end{theorem}

\begin{equation}
     G=\Delta  +\sqrt{\frac{2R^2}{n}}\left(\sqrt{I(x^+,y)}-\sqrt{I(x,y)}\right) < 0
\end{equation}

where $\Delta = \mathcal{E}_{\hat{\mathbb{P}_+}}-\mathcal{E}_{\hat{\mathbb{P}}}$. The proof is given in the Appendix. Following \textbf{Theorem} \ref{app:the1}, it is evident that incorporating label-related features/sensors (i.e., $a$) can effectively reduce the generalization error bound, thereby enhancing the generalization performance.

\noindent
\textbf{Proof}:
For $\mathcal{E}_{\hat{\mathbb{P}}}\left(f\right)$ and $\mathcal{E}_{\hat{\mathbb{P}}_+}\left(f\right)$, we have:
\begin{equation}
   \begin{aligned}
\mathcal{E}&_{\hat{\mathbb{P}}}\left(f\right)=\mathcal{L}(f(x),y)\\
\mathcal{E}&_{\hat{\mathbb{P}}_+}\left(f\right)=\mathcal{L}(f(\left[x,a\right]),y)\\
   \end{aligned}
\end{equation}
note that the  $x,a$ are conditionally independent given  $y$, thus, we can easily have:
\begin{equation}
    \label{eq2}
    \Delta=\mathcal{E}_{\hat{\mathbb{P}}_+}-\mathcal{E}_{\hat{\mathbb{P}}}\leq 0
\end{equation}

For $\sqrt{\frac{2R^2}{n}}\left(\sqrt{I(x^+,y)}-\sqrt{I(x,y)}\right)$, since $\sqrt{\frac{2R^2}{n}}>0$, $I(x^+,y)\geq 0$ and $I(x,y)\geq 0$. Thus,  The positivity or negativity of the expression can be determined by $I(x^+,y)-I(x,y)$. Then we have:
\begin{equation}
\label{eq4}
    \begin{aligned}
        I(x^+,y)-I(x,y)
        &=H(Y)-H(Y|x^+)\\
        &\quad-\left(H(Y)-H(Y|x)\right)\\
        &=H(Y|x)-H(Y|x^+)
      \end{aligned}
\end{equation}
where $H(\cdot)$ is the entropy, since $x^+=[x,a]$,  $x_l,x_{\setminus l}$ are independent of each other, $x,a$ are depend of $y$, $H(Y|x^+)-H(Y|x) < 0$, then we have:
\begin{equation}
\label{eq5}
    \begin{aligned}
        I(x^+,y)-I(x,y)=H(Y|x)-H(Y|x^+)< 0
      \end{aligned}
\end{equation}
According to Eq. \ref{eq5}, we easily have:
\begin{equation}
\label{eq7}
    \sqrt{\frac{2R^2}{n}}\left(\sqrt{I(x^+,y)}-\sqrt{I(x,y)}\right) < 0
\end{equation}
Combining Eqs \ref{eq2} and \ref{eq7}, we have:

\begin{equation}
    G=\Delta  +\sqrt{\frac{2R^2}{n}}\left(\sqrt{I(x^+,y)}-\sqrt{I(x,y)}\right) < 0
\end{equation}
Thus, with the incorporates of the new sensors, the $\mathcal{E}_{\mathbb{P}}$  have a tighten upper bound.

\noindent
\textbf{Q.E.D}

\begin{theorem}
\label{app:the2}
Given a random variable $\hat{a}$ generated from a distribution $\mathbb{Q}_{\hat{a}}$, and $\hat{x}^+=[x, \hat{a}]\sim \mathbb{Q}$ , then we have:
\end{theorem}
\begin{equation}
\begin{aligned}
    \mathcal{E}_{\mathbb{P}_+}\left(f\right)
    &\leq \mathcal{E}_{\hat{\mathbb{Q}}}\left(f\right)
    +\epsilon_{ideal}
    +d_{\mathcal{H}\Delta\mathcal{H}}(\mathbb{Q}_{a},\mathbb{Q}_{\hat{a}})\\
    &\quad+\sqrt{\frac{2R^2}{n}I(x,y)}
\end{aligned}
\end{equation}

where $\mathcal{E}_{\hat{\mathbb{Q}}}\left(f\right)$ is the empirical generalization error over distribution $\mathbb{Q}$, $d_{\mathcal{H}\Delta\mathcal{H}}(\mathbb{Q}_{a},\mathbb{Q}_{\hat{a}})$ is the $\mathcal{H}\Delta\mathcal{H}$ distance between $\mathbb{Q}_{a}$ and $\mathbb{Q}_{\hat{a}}$.

Before we proof the \textbf{Theorem} \ref{app:the2}, we need the following Definition:
\begin{definition}[Disparity]
    Given two classifier $f$ and $f'$, the disparity between two classifier $f$ and $f'$ over the distribution $\mathbb{P}$ is as follows:
\end{definition}
\begin{equation}
    \mathcal{E}_\mathbb{P}(f,f')_=\mathbb{E}_{x\sim\mathbb{P}}\left[f(x)\neq f'(x)\right]
\end{equation}
\begin{definition}[Ideal Classifier]
    Given two Distribution $\mathbb{P}$and $\mathbb{Q}$ the ideal classifier $f^*$ have the minimum risk over two distributions as follows:
\end{definition}
\begin{equation}
    f^*=\arg\min_f\left(\mathcal{E}_\mathbb{P}(f)+\mathcal{E}_\mathbb{P}(f)\right)
\end{equation}
\begin{definition}[$\mathcal{H}\Delta\mathcal{H}$ distance]
     Given two Distribution $\mathbb{P}$ and $\mathbb{Q}$, the $\mathcal{H}\Delta\mathcal{H}$ distance between $\mathbb{P}$ and $\mathbb{Q}$ is defined as follows:
\end{definition}
\begin{equation}
    d_{\mathcal{H}\Delta\mathcal{H}}(\mathbb{P},\mathbb{Q})=\sup_{f,f'\in\mathcal{H}} |\mathcal{E}_\mathbb{P}(f,f')-\mathcal{E}_\mathbb{Q}(f,f')|
\end{equation}
where $\mathcal{H}$ is the Hypothesis space.

\noindent
\textbf{Assumption:} the ideal classifier has a small risk.
\begin{equation}
    \epsilon_{ideal}=\left(\mathcal{E}_\mathbb{P}(f^*)+\mathcal{E}_\mathbb{P}(f^*)\right)
\end{equation}
\noindent
\textbf{Proof:}

By using the triangle inequalities, we have
\begin{equation}
    \begin{aligned}
    \mathcal{E}_{\mathbb{P}_+}\left(f\right)
    &\leq \mathcal{E}_{\mathbb{P}_+}\left(f^*\right)
    +\mathcal{E}_{\mathbb{P}_+}\left(f,f^*\right)\\
    &\leq \mathcal{E}_{\mathbb{P}_+}\left(f^*\right)
    +\mathcal{E}_{\mathbb{Q}}\left(f,f^*\right)\\
    &\quad+\mathcal{E}_{\mathbb{P}_+}\left(f,f^*\right)
    -\mathcal{E}_{\mathbb{Q}}\left(f,f^*\right)\\
    &\leq \mathcal{E}_{\mathbb{P}_+}\left(f^*\right)
    +\mathcal{E}_{\mathbb{Q}}\left(f,f^*\right)\\
    &\quad+\left|\mathcal{E}_{\mathbb{P}_+}\left(f,f^*\right)
    -\mathcal{E}_{\mathbb{Q}}\left(f,f^*\right)\right|\\
    &\leq \mathcal{E}_{\mathbb{Q}}\left(f\right)
    +\mathcal{E}_{\mathbb{Q}}\left(f^*\right)
    +\mathcal{E}_{\mathbb{P}_+}\left(f^*\right)\\
    &\quad+\left|\mathcal{E}_{\mathbb{P}_+}\left(f,f^*\right)
    -\mathcal{E}_{\mathbb{Q}}\left(f,f^*\right)\right|
    \end{aligned}
\end{equation}

According to definition of ideal calssifier and $\mathcal{H}\Delta\mathcal{H}$ disatance we have:
\begin{equation}
\label{eq15}
    \begin{aligned}
    \mathcal{E}_{\mathbb{P}_+}\left(f\right)
    &\leq \mathcal{E}_{\mathbb{Q}}\left(f\right)
    +\mathcal{E}_{\mathbb{Q}}\left(f^*\right)
    +\mathcal{E}_{\mathbb{P}_+}\left(f^*\right)\\
    &\quad+\left|\mathcal{E}_{\mathbb{P}_+}\left(f,f^*\right)
    -\mathcal{E}_{\mathbb{Q}}\left(f,f^*\right)\right|\\
    &= \mathcal{E}_{\mathbb{Q}}\left(f\right)
    +\epsilon_{ideal}
    +\left|\mathcal{E}_{\mathbb{P}_+}\left(f,f^*\right)
    -\mathcal{E}_{\mathbb{Q}}\left(f,f^*\right)\right|\\
    &\leq \mathcal{E}_{\mathbb{Q}}\left(f\right)
    +\epsilon_{ideal}
    +d_{\mathcal{H}\Delta\mathcal{H}}(\mathbb{P}_+,\mathbb{Q})
    \end{aligned}
\end{equation}

According to Lemma \ref{lemm2}, we have:
\begin{equation}
    \mathcal{E}_{\mathbb{P}_+}\left(f\right)\leq\mathcal{E}_{\hat{\mathbb{P}}_+}\left(f\right)+\sqrt{\frac{2R^2}{n} I(x,y)}
\end{equation}
Note that $\hat{x}^+=[x,\hat{a}]$ and $x^+=[x,a]$, we further have:
\begin{equation}
\label{eq17}
\begin{aligned}
        d_{\mathcal{H}\Delta\mathcal{H}}(\mathbb{P}_+,\mathbb{Q})
        &\leq d_{\mathcal{H}\Delta\mathcal{H}}(\mathbb{P},\mathbb{P})\\
        &\quad+d_{\mathcal{H}\Delta\mathcal{H}}(\mathbb{Q}_{a},\mathbb{Q}_{\hat{a}})\\
        &\leq d_{\mathcal{H}\Delta\mathcal{H}}(\mathbb{Q}_{a},\mathbb{Q}_{\hat{a}})
\end{aligned}
\end{equation}
By plugging Lemma 1 and Eq.\ref{eq17} into Eq. \ref{eq15}, we have:
\begin{equation}
\begin{aligned}
    \mathcal{E}_{\mathbb{P}}\left(f\right)
    &\leq \mathcal{E}_{\hat{\mathbb{Q}}}\left(f\right)
    +\epsilon_{ideal}
    +d_{\mathcal{H}\Delta\mathcal{H}}(\mathbb{Q}_{a},\mathbb{Q}_{\hat{a}})\\
    &\quad+\sqrt{\frac{2R^2}{n} I(x,y)}
\end{aligned}
\end{equation}

\textbf{Q.E.D}

\begin{lemma}\cite{mansour2009domain}
\label{lemma3}
    Let $S$ and $T$ be the source and target domains over $\mathcal{X} \times \mathcal{Y}$, respectively. Let $\mathcal{H}$ be a hypothesis class, and let $\ell : \mathcal{Y} \times \mathcal{Y} \rightarrow \mathbb{R}_+$ be a loss function that is symmetric, obeys the triangle inequality, and is bounded, $\forall (y, y') \in \mathcal{Y}^2$, $\ell(y, y') \leq M$ for some $M > 0$. Then, for $h_S^* = \arg\min\limits_{h \in \mathcal{H}} R_S^\ell(h)$ and $h_T^* = \arg\min\limits_{h \in \mathcal{H}} R_T^\ell(h)$ denoting the ideal hypotheses for the source and target domains, we have

\end{lemma}

\begin{equation}
    \forall h \in \mathcal{H}, \, R_T^\ell(h) \leq R_S^\ell(h, h_S^*) + d_{\mathcal{H}\Delta\mathcal{H}}(\mathbb{P}_S, \mathbb{P}_T) + \epsilon,
\end{equation}

where $R_S^\ell(h, h_S^*) = \mathbb{E}_{x \sim S_X} \ell(h(x), h_S^*(x))$ and $\epsilon = R_T^\ell(h_T^*) + R_S^\ell(h_T^*, h_S^*)$.
\begin{theorem}
\label{app:the3}
 Let $\mathcal{H}$ be a hypothesis class. $R_s^\ell(g)$ and $R_T^\ell(g)$  denote the expected loss of loss function $\ell$  for the source and target domains. for $g_s^* = \arg\min\limits_{g \in \mathcal{H}} R_s^\ell(g)$ and $g_T^* = \arg\min\limits_{h \in \mathcal{H}} R_T^\ell(g)$ denoting the ideal hypotheses for the source and target domains, we have
\end{theorem}
\begin{equation}
 \, R_T^\ell(g_s) \leq R_S^\ell(g_s, g_S^*) + d_{\mathcal{H}\Delta\mathcal{H}}(\mathbb{P}_T^\cap,\mathbb{P}_s^\cap) + \epsilon,
\end{equation}
where $R_S^\ell(g_s, g_S^*) = \mathbb{E}_{x \sim \mathbb{P}_s} \ell(g_s(x), g_s^*(x))$ and $\epsilon = R_T^\ell(g_T^*) + R_S^\ell(g_T^*, g_s^*)$.

\noindent
\textbf{Proof:}
The proof of Theorem \ref{app:the3} can be obtained by substituting the variables in Lemma \ref{lemma3}, so the proof is omitted.
\vfill

\vfill

\end{document}